\documentclass{llncs}
\usepackage[pdftex]{graphicx}
\usepackage{subfig}
\usepackage{hyperref}
\usepackage[font=small,labelfont=bf]{caption}
\begin{document}

\title{To Learn or Not to Learn Features for Deformable Registration?}
\titlerunning{To Learn or Not to Learn?}  
%
\author{Aabhas Majumdar, Raghav Mehta And Jayanthi Sivaswamy}
%
%

\institute{Center for Visual Information Technology (CVIT), IIIT-Hyderabad, India}

\maketitle              

\begin{abstract}
Feature-based registration has been popular with a variety of features ranging from voxel intensity to Self-Similarity Context (SSC). In this paper, we examine the question of how features learnt using various Deep Learning (DL) frameworks can be used for deformable registration and whether this feature learning is  necessary or not. We investigate the use of features learned by different DL methods in the current state-of-the-art discrete registration framework and analyze its performance on 2 publicly available datasets. We draw insights about the type of DL framework useful for feature learning. We consider the impact, if any, of the complexity of different DL models and brain parcellation methods on the performance of discrete registration. Our results indicate that the registration performance with DL features and SSC are comparable and stable across datasets whereas this does not hold for low level features. This shows that when handcrafted features are designed based on good insights into the problem at hand, they perform better or are comparable to features learnt using deep learning framework.
\end{abstract}

\section{Introduction}

Deformable image registration is critical to tasks such as surgical planning, image fusion, disease monitoring etc.\cite{ImageFusion}. We focus on application to neuro images where tasks such as Multi-atlas segmentation \cite{MAS} and atlas construction \cite{Atlas},  require registration to handle variations in the shape and size of brain across subjects. Registration entails minimizing a cost function through iterative optimization. Since the cost function quantifies the similarity between the two images to be registered, it plays a crucial role in determining the accuracy of results. Traditional image intensity-based approaches define cost functions based on mutual information, sum of squared difference etc., \cite{SimiMeasu} and use continuous optimization to find the required deformation field. Continuous optimization based methods require cost functions to be differentiable. With a discrete optimization (DO) formulation, registration has been shown \cite{DO} \cite{FEM} to be more efficient with a 40 to 50-fold reduction in computational time, no loss in accuracy and no requirement of differentiability for cost function. This allows them to use simple cost function like Sum of Absolute Difference (SAD).

Feature based registration was proposed \cite{HAMMER} as an improvement over the intensity-based approach. Features that have been explored range from normalized intensity values, edges, geometric moments \cite{HAMMER}, 3D Gabor attributes \cite{DRAMMS} and a Modality Independent Neighborhood Descriptor \cite{MIND}. The natural question to ask is if it is better to learn the features, instead of using hand-crafted ones, since the experience of learning features (using deep networks) for another important problem, namely, segmentation, has been positive \cite{AnatoBrain}. Deep features learnt using unsupervised method \cite{UHAMMER}, has been shown to perform better than traditional features like intensity and edges. A Co-Registration and Co-Segmentation framework \cite{CoRegCoSeg} has also been proposed using learnt priors on 8 sub-cortical structures and those learnt using a Convolutional Neural Network (CNN) is reported to outperform those learnt with Random-Forrest based classifiers. While these few reports indicate the potential benefit of deep feature learning for registration, it is of interest to gain deeper insights into this specific approach, given the well-established high cost, particularly training overhead, of deep learning which may deter clinical applications with this approach.

In this paper, we seek to gain insights by delving deeper into the issue of feature learning for registration. We attempt to answer the following six questions regarding deep feature learning through extensive experiments on two publicly available datasets. (i) Does complexity of learning architecture matter? (ii) What kind of learning strategy is useful? Supervised or Unsupervised? (iii) What features are better in supervised feature learning? (iv) Does registration accuracy vary with the number of labeled structures in the  training data? (v) Does difference in parcellation during training matter? and finally the main question: (vi) To learn or not to learn features for deformable image registration?

To obtain answers to the above questions, a testbed was created for registration using the discrete optimisation framework described in \cite{wbir} and different Deep Neural Networks (DNNs) were considered for learning the features. The suitability of these features for registration was assessed in terms of the Jaccard Coefficient. At the end, comparison was done against standard low level and high level features like intensity and Self-Similarity Context \cite{SSC} to assess the requirement of feature learning using deep learning methods.

\vspace{-3mm}
\section{Method}

This is a brief overview of the two frameworks which we adopted to setup our testbed and experiments: Discrete Optimization (DO) based registration and Deep Learning based Feature Learning. 

\vspace{-2mm}
\subsection{Discrete Optimization}

Given a fixed image $I_f$ and moving image $I_m$ the deformation field $u$ required to align $I_m$ to $I_f$ is found via optimizing a cost function $E(u)$. The DO method presented in \cite{wbir} determines $u$ by minimizing $E(u)$:

\begin{equation}
 E(u) = \sum_{\Omega}  S(I_f, I_m, u) + \alpha |\nabla u|^2
\end{equation}

where $\Omega$ is the image patch. The first term $S$ denotes the similarity function between the fixed and warped images while the second term is the regularization term. In our experiments, SAD was chosen as $S$ for the ease of computation.

In DO, the deformation field is only allowed to take values from a quantized set of 3-D displacement for each voxel $x$. $d \in   \{0,\pm q,\pm 2q, ..., \pm l_{max} \}$ Here, $q$ is the quantization step and $l_{max}$ is the maximum range of displacement.

A six-dimensional displacement space volume is created whose each entry is the point-wise similarity cost of translating a voxel $x$ with a displacement $d$:
\vspace{-2mm}
\begin{equation}
DSV(x,d) = S(I_f(x), I_m(x+d))
\end{equation} 

Here, $I_f(x)$ can be simply the voxel intensity or a feature representing the voxel. The displacement field is obtained by winner-takes-all method by selecting the field with the lowest cost for each voxel: $ u =  \arg \min_d(DSV(d))$. For further, detail into this framework, we refer readers to \cite{wbir}.

\vspace{-3mm}
\subsection{Deep Learning Framework}

DNNs are inspired by the biological networks akin to the multilayer perceptron. Basic blocks of DNNs are convolutional layer (2D/3D), maxpooling layer (2D/3D), fully connected layer, dropout layer, activation functions like Rectified Linear Unit (ReLU), tanh, softmax, and batch normalization layer. For a detailed description of these blocks readers are referred to \cite{AlexNet}.

\begin{figure*}[!t]
\centering
\includegraphics[scale=0.145]{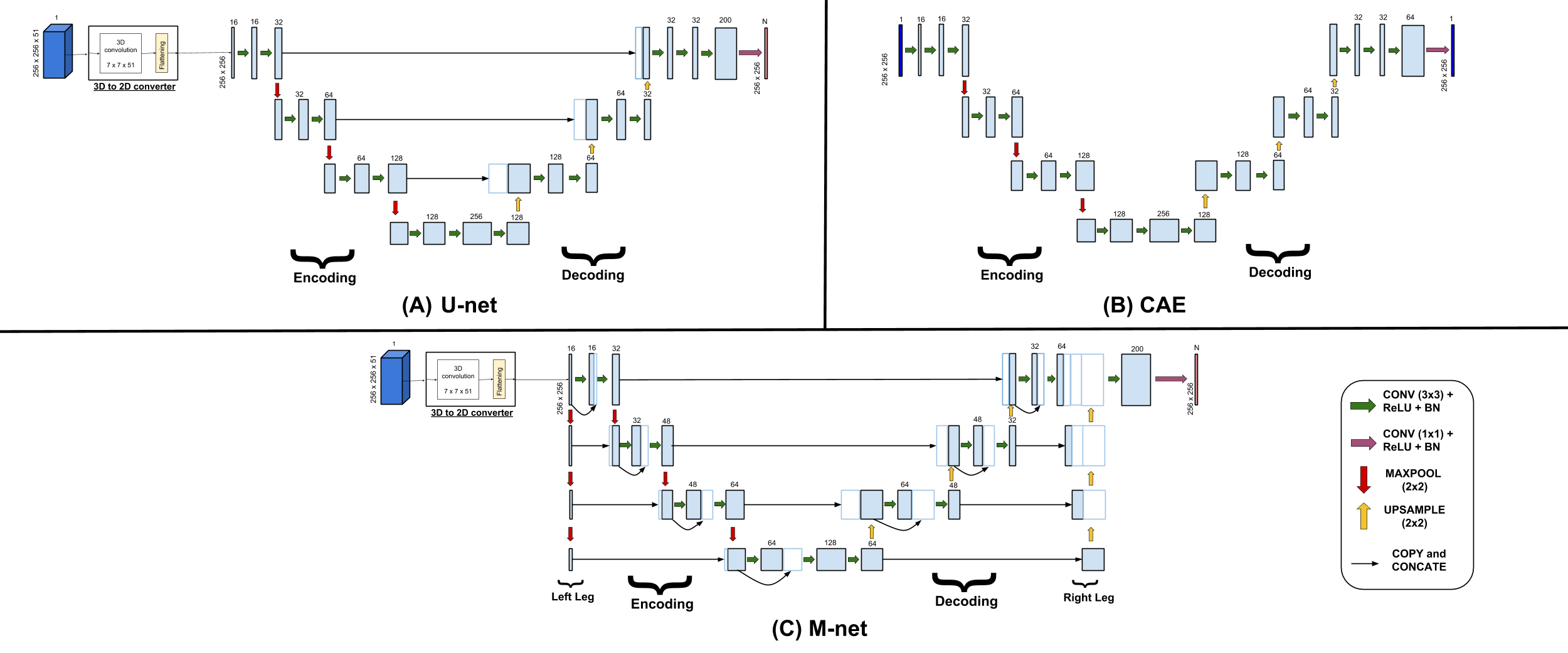}
\center \caption{Three DNNs used for feature learning}
\label{fig:Deep_Architectures}
\end{figure*}

A Deep Learning framework can be broadly of two types: (i) Supervised and (ii) Unsupervised. The main difference between these two types is that in the former, for any input $X$, the network tries to predict output $Y$, which is a class label, while in the latter, the network tries to predict $X$ using the same $X$ as input and in this way network learns something intrinsic about the data without the help of labels generally created by humans.

A CNN \cite{AlexNet} is a widely used DNN for supervised learning. CNNs have been used for various tasks such as segmentation and classification. Similarly, Convolutional AutoEncoder (CAE) \cite{CAE} is a popular framework for unsupervised learning. They are used for varied type of tasks such as learning hidden (or lower dimensional) representation of data, denoising etc.

In our experiments, two CNN architectures, namely U-net \cite{Unet} and M-net \cite{M-net} were used for supervised learning. The M-net is an improvement on U-net with added residual and supervision connections. A stack of slices (51) as 3D input is passed through 3D-to-2D converted and then processed by both the architectures to produce segmentation for center slice, as shown in Fig:\ref{fig:Deep_Architectures} A and C. For more details about these architectures we refer readers to \cite{Unet} and \cite{M-net}.

CAE is also a DNN which learns useful lower dimensional representation of input from which original input can be generated back with minimal loss of information. The CAE architecture used in our experiment is shown in Fig:\ref{fig:Deep_Architectures}(B).

\vspace{-3mm}
\section{Experiments and Results:}
\subsection{Datasets Description}
\begin{figure*}[!t]
\centering
\includegraphics[width = 3.8in]{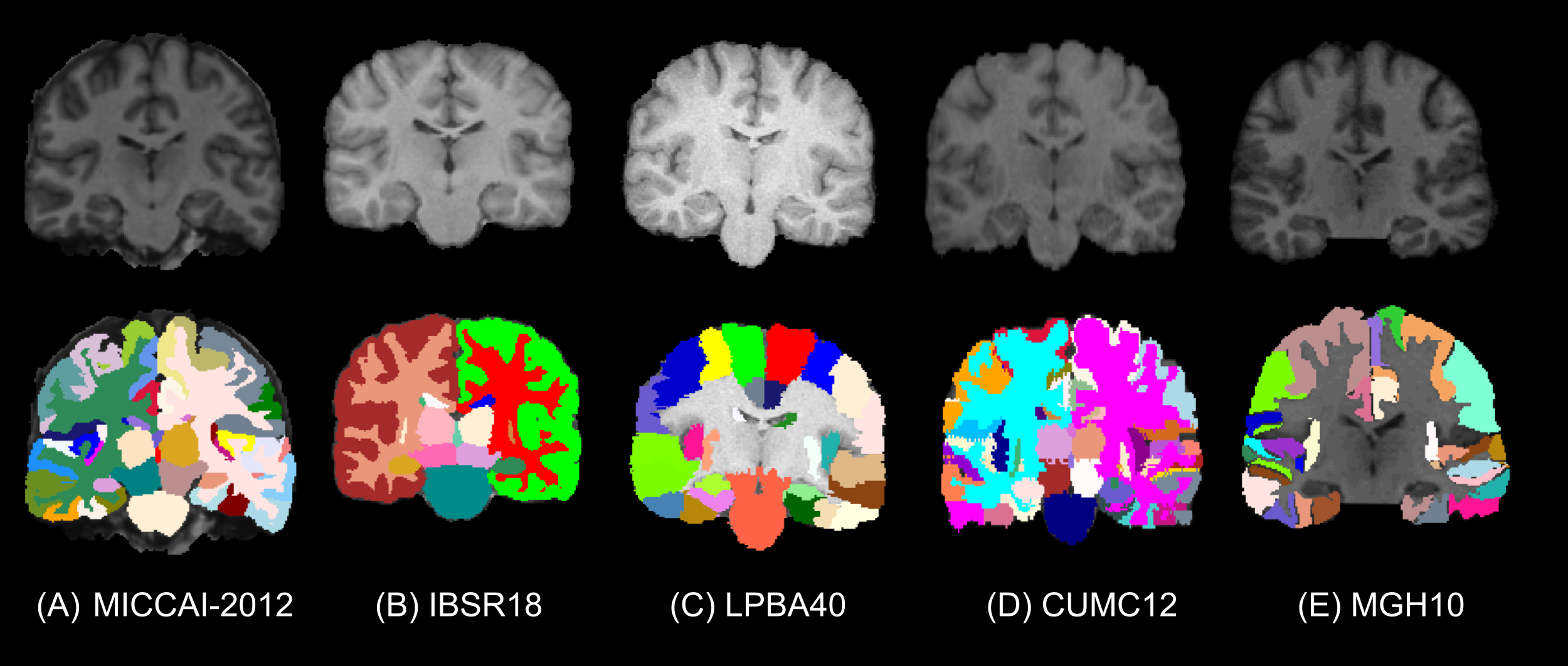}
\caption{Coronal slices of brain images (Top Row) and their manual segmentation (Bottom Row) of DL training (MICCAI-2012, IBSR18 and LPBA40) and registration testing (CUMC12 and MGH10) datasets.}
\label{Dataset_Slices}
\end{figure*}

In order to ensure thorough evaluation, different datasets were employed for training (the DNNs) versus testing (the registration module).  Sample slices of all these datasets are shown in Fig:\ref{Dataset_Slices}.

\vspace*{-10mm}
\begin{table}
\centering
\caption{Deep Learning Training Dataset Description}
\label{training datasets}
\resizebox{\textwidth}{!}{%
\begin{tabular}{|c|c|c|c|c|}
\hline
Dataset     & \# Training Volumes & \# Validation Volumes & \# labels & Parcellation Type                       \\ \hline
MICCAI-2012 & 15                  & 20                    & 135       & Whole Brain (Cortical and Non-cortical) \\ \hline
IBSR18      & 15                  & 3                     & 32        & Whole Brain (Cortical and Non-cortical) \\ \hline
LPBA40      & 30                  & 10                    & 57        & Partial Brain (Mainly Cortical)         \\ \hline
\end{tabular}}
\vspace{-5mm}
\end{table}

\textit{Deep Learning Training Datasets:} Datasets used for training were chosen according to the diversity in total number of labeled structures and the structure parcellation methods. Details of the chosen datasets are given in Table:\ref{training datasets} 

\textit{Registration Testing Datasets:} The datasets for testing the registration accuracy were chosen based on their popularity for evaluating registration \cite{14RegComp} and variation in structure parcellation methods.  \textit{(i) CUMC12:} This dataset has 12 MRI volumes, which are manually labeled into 130 structures. \textit{(ii) MGH10:} This dataset has 10 volumes, segmented into 106 structures. It should be noted that unlike CUMC12 dataset, only cortical structures are marked for this dataset.

\vspace{-3mm}
\subsection{Evaluation Metric:}
Registration performance was evaluated using mean Jaccard Coefficient (JC). This is the standard evaluation metric employed for comparison of 14 Registration methods in \cite{14RegComp}. JC between two binary segmentation A and B is defined as: $JC(A,B) = \frac{|A \cap B|}{|A \cup B|} * 100$. Throughout this paper, we compare registration performance with mean JC which is computed as follows: JC is averaged first across $N$ individual structures for a single volume; then it is averaged across $M$ pairwise registration output. Thus, to evaluate registration performance on the CUMC12 dataset, average JC is found over $N=$ 130 structures in a volume and then the average JC is computed for $M=$ 144 pairwise registration outputs.

\vspace{-3mm}
\subsection{Implementation Detail:} 
All the DNNs were trained on a NVIDIA K40 GPU, with 12GB of RAM for 30 epochs. Approximate training time was 3 days. The CNN was trained using Adam Optimizer with following hyper parameters: LR=0.001, $\beta_1$=0.9, $\beta_2$=0.99, and $\epsilon=10*e^{\neg 8}$. LR was reduced by a factor of 10 after 20 epochs. Code was written in Keras Library using Python. The C++ code for DO-based registration made publicly available by the authors of \cite{wbir} was used. The python code for Deep Learning was integrated in C++ for a seamless implementation. 

The effect of intensity variation among training and testing datasets was handled by matching the intensity of all the volumes of testing datasets to that of training dataset volume using Intensity Standardization (IS) \cite{IS}.

\vspace{-3mm}
\subsection{Feature Learning Experiments and Results:}
\begin{figure*}[!t]
\centering
\subfloat[]{\includegraphics[width = 2.4in]{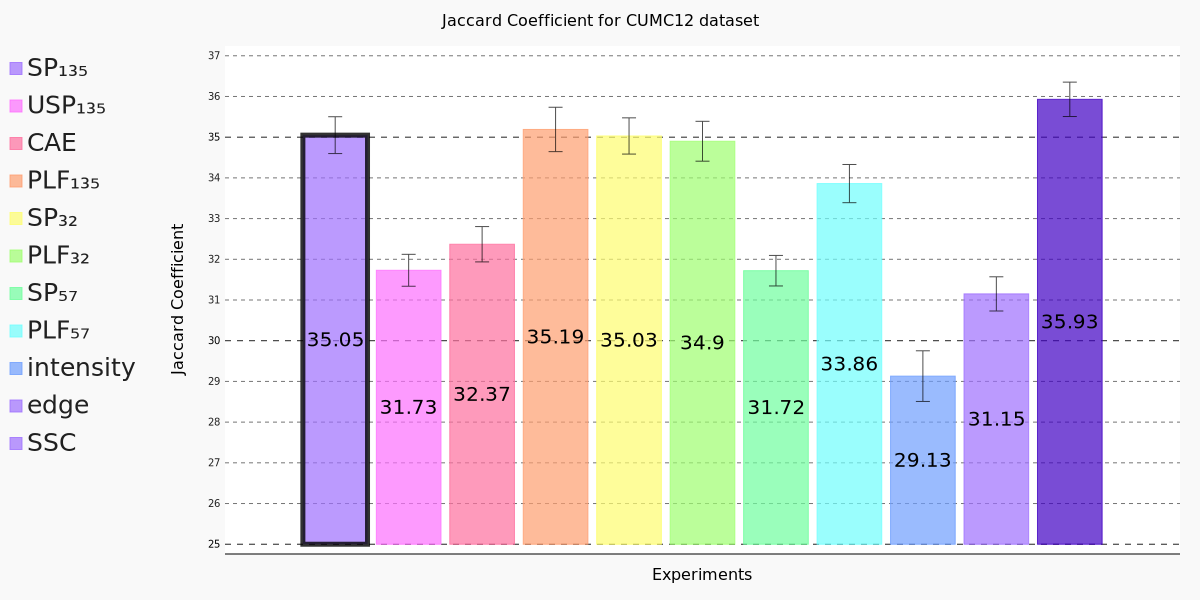}}
\subfloat[]{\includegraphics[width = 2.4in]{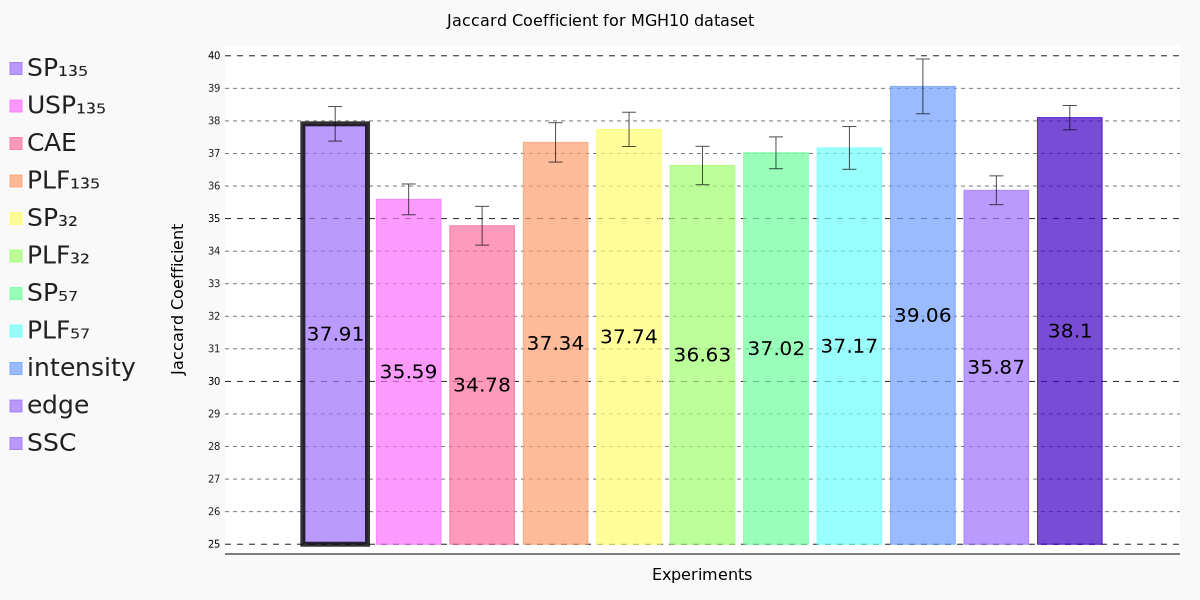}}
\caption{Registration performance comparison, in terms of JC, for various deep features on (a) CUMC12 dataset and (b) MGH10 dataset. Here, SP denotes Segmentation Priors and PLF denotes Penultimate Layer Features.}
\label{fig:graphs}
\end{figure*}

A set of experiments were performed to gain insights into the following six issues in the context of feature-based deformable registration. Registration performance of all these experiments in terms of mean JC is given in Fig:\ref{fig:graphs}.

\textbf{\textit{Role of complexity of learning architecture:}} Both the U-net and M-net were trained on the MICCAI-2012 dataset and the Segmentation Priors (SP) features  were extracted from U-net (USP$_{135}$) and M-net (SP$_{135}$) for registration. The mean JC obtained for USP$_{135}$ and SP$_{135}$ on the MGH10 dataset were $35.59$ and $37.90$ respectively, while they were $31.73$ and $35.05$ for the CUMC12 dataset. These results indicate that the complexity of architecture does play an important role in feature extraction as registration performance is better for features extracted from a more complex network (M-net).

\textbf{\textit{Supervised or Unsupervised learning of features? :}} The M-net and CAE were trained on the MICCAI-2012 dataset. The SP features were extracted from the M-net (SP$_{135}$) and the Penultimate Layer Features (PLF) from CAE. The mean JC obtained for SP$_{135}$ and CAE were $37.90$ and $34.77$, respectively for the MGH10 dataset while they were $35.05$ and $32.37$ for the CUMC12 dataset. Thus, the supervised feature learning appears to be more effective. 

\textbf{\textit{Choice of learnt features:}} Registration was done with SP and PLF (hidden layer representation) features separately, after training M-net on the MICCAI-2012 dataset. The obtained mean JC, as shown in Fig:\ref{fig:graphs}, indicates that SP and PLF give comparable performance across both datasets (CUMC12: SP$_{135}$ = 35.05 and PLF$_{135}$ = 35.19; MGH10: SP$_{135}$= 37.91 and PLF$_{135}$ = 37.34). 

\textbf{\textit{Role of the number of labeled structures in training data:}} Available training datasets vary in terms of the number of labeled structures. We can expect the feature learnt on dataset with more structure to differentiate between its neighbouring structures in a better way. In order to understand how this can impact registration, the M-net was trained on two different datasets, namely, MICCAI-2012 (labels: 135) and IBSR18 (labels: 32). The SP and PLF features were extracted from the CNN (M-net) and used in registration. The obtained mean JC for both SP and PLF on both CUMC12 and MGH10 were comparable. (CUMC12: SP$_{135}$ = 35.05, SP$_{32}$ = 35.03, PLF$_{135}$ = 35.19 and PLF$_{32}$ = 34.9; MGH10: SP$_{135}$ = 37.9, SP$_{32}$ = 37.73, PLF$_{135}$ = 37.34 and PLF$_{32}$ = 36.63) Thus, the features learnt with different number of labeled structures appear to be equally effective for registration. A possible reason for this can be that MICCAI-2012 and IBSR18 datasets have equal number of labels for sub-cortical structures and white matter. However, the former set has a finer level parcellation for cortical structures which essentially encodes spatial position and local information and this may not give added advantage over coarser level parcellation for registration, as registration inherently encodes this information.

\textbf{\textit{Parcellation of training dataset:}} Fig:\ref{Dataset_Slices} shows that while the whole brain is marked in MICCAI-2012 and IBSR18 datasets, only cortical structures are marked in LPBA40. In order to assess the effect of various parcellation methods, the M-net was trained on the LPBA40 and MICCAI-2012 datasets. SP (SP$_{57}$, SP$_{135}$) and PLF (PLF$_{57}$, PLF$_{135}$) features were extracted from both. The registration accuracy for both CUMC12 and MGH10 datasets are shown in Fig:\ref{fig:graphs}. It can be seen that there is a drop in JC of approximately $3.33$ (9.5\%) and $1.33$ (3.8\%) for SP and PLF respectively, on CUMC12 dataset, relative to the value obtained with features from M-net trained on the MICCAI-2012 dataset (SP$_{135}$ = 35.05, SP$_{57}$ = 31.72 and PLF$_{135}$ = 35.19, PLF$_{57}$ = 33.86); whereas on MGH10 dataset, there is only marginal drop in JC of $0.89$ (2.3\%) and $0.18$ (0.4\%) for SP and PLF, respectively (SP$_{135}$ = 37.90, SP$_{57}$ = 37.01 and PLF$_{135}$ = 37.34, PLF$_{57}$ = 37.16). This can be attributed to the fact that both LPBA40 and MGH10 have only cortical structures marked, while CUMC12 has both cortical and non-cortical structures. Overall, the above results suggest that parcellation method of training dataset should be an important consideration in feature-based registration. Further, it is advisable to train a CNN on a dataset with parcellation for both cortical and sub-cortical structures.

\textit{\textbf{To learn or not to learn features for deformable image registration? }} Finally, we turn to the main question of interest: the necessity of feature learning. The registration accuracy of features learnt using M-net was compared against low level features such as intensity, edges as well as a higher level feature, namely, Self-Similarity Context (SSC). The JC values obtained are shown in Fig:\ref{fig:graphs}. Raw intensity feature with SAD as similarity metric has the best performance on MGH10 dataset (39.05) but not on the CUMC12 (29.13) dataset. This is most likely to be due to the persistent voxel intensity variation between the datasets (MGH10 has 32 and CUMC12 has 512 distinct values) despite IS. Interestingly, while both learnt (SP) and high level (SSC) features yield more robust performance across datasets, the latter performs marginally better (CUMC12: SSC = 35.93 and SP$_{135}$ = 35.05; MGH12: SSC = 38.1 and SP$_{135}$ = 37.9). Taking the mean JC difference between CUMC12 and MGH10 as a quantifier of robustness, the obtained results (2.84(SP$_{135}$), 2.17 (SSC), 4.72 (edge) and 9.93 (intensity)), indicate that learning \textit{may not} give results superior to hand-crafting of features. SSC is a feature explicitly derived for registration whereas learnt features such as SP are optimised for good segmentation as they are trained on a segmentation dataset.


\vspace{-5mm}
\section{Conclusions}

In this paper, the issue of employing learnt (with DNN) features for deformable registration was explored in considerable detail with a set of experiments. Some of the experimental findings such as superiority of supervised features over unsupervised features in terms of robustness is intuitive while others such as accuracy being insensitive to change in the total number of labeled structures during supervised training are counter-intuitive. Our methodology for learning features from a segmentation network was motivated by the widespread practice of assessing registration accuracy indirectly via segmentation as the latter has well defined evaluation metrics. This approach is attractive when \textit{both} problems need to be solved. Learning features (which leads to robust, yet marginally lower performance than SSC) requires considerable computational resources, as one pairwise registration takes 2 mins of CPU + 8 mins of GPU time for feature learnt with DNNs, while it only takes 2-3 mins on CPU for SSC. Recent papers \cite{quicksilver},\cite{SVFnet} have  tried to directly learn deformation field for registration instead of features and \cite{quicksilver} appears to have slightly better performance than SSC.  Taking our findings and based on recent reports, SSC may be a better option in low-resource settings and limited annotated data scenario, especially, if only registration is of interest. 




\begin{thebibliography}{3}
\footnotesize
\bibitem{ImageFusion}
De Nigris, D., et al., 2013. Fast rigid registration of pre-operative magnetic resonance images to intra-operative ultrasound for neurosurgery based on high confidence gradient orientations. IJCARS, 8(4), pp.649-661.

\bibitem{MAS}
Iglesias, J.E. and Sabuncu, M.R., 2015. Multi-atlas segmentation of biomedical images: a survey. MedIA, 24(1), pp.205-219.

\bibitem{Atlas}
Joshi, S., et al., 2004. Unbiased diffeomorphic atlas construction for computational anatomy. NeuroImage, 23, pp.S151-S160.

\bibitem{SimiMeasu}
Penney, G.P., et al., 1998. A comparison of similarity measures for use in 2-D-3-D medical image registration. IEEE TMI, 17(4), pp.586-595.

\bibitem{DO}
Heinrich, M.P., et al., 2013. Uncertainty estimates for improved accuracy of registration-based segmentation propagation using discrete optimisation. In MICCAI Challenge Workshop on Segmentation: Algorithms, Theory and Applications.

\bibitem{SSC}
Heinrich, M.P., et al., 2013. Towards realtime multimodal fusion for image-guided interventions using self-similarities. In MICCAI (pp. 187-194). 

\bibitem{wbir}
Heinrich, M.P., et al., 2014. Non-parametric discrete registration with convex optimisation. In WBIR (pp. 51-61). 

\bibitem{FEM}
Popuri, K., et al., 2013. A variational formulation for discrete registration. In MICCAI (pp. 187-194). 

\bibitem{MIND}
Heinrich, M.P., et al., 2012. MIND: Modality independent neighbourhood descriptor for multi-modal deformable registration. MedIA, 16(7), pp.1423-1435.

\bibitem{14RegComp}
Klein, A., et al., 2009. Evaluation of 14 nonlinear deformation algorithms applied to human brain MRI registration. Neuroimage, 46(3), pp.786-802.

\bibitem{HAMMER}
Shen, D. and Davatzikos, C., 2002. HAMMER: hierarchical attribute matching mechanism for elastic registration. IEEE TMI, 21(11), pp.1421-1439.

\bibitem{DRAMMS}
Ou, Y., et al., 2011. DRAMMS: Deformable registration via attribute matching and mutual-saliency weighting. MedIA, 15(4), pp.622-639.

\bibitem{CoRegCoSeg}
Shakeri, M., et al., 2016. Prior-based Coregistration and Cosegmentation. In MICCAI (pp. 529-537). 

\bibitem{UHAMMER}
Wu, G., et al., 2016. Scalable High-Performance Image Registration Framework by Unsupervised Deep Feature Representations Learning. IEEE TBME, 63(7), pp.1505-1516.

\bibitem{quicksilver} Yang, X., et al., 2017. Quicksilver: Fast predictive image registration-A deep learning approach. NeuroImage, 158, pp.378-396.


\bibitem{SVFnet} Rohé, M.M., et al., 2017. SVF-Net: Learning Deformable Image Registration Using Shape Matching. In MICCAI (pp. 266-274).


\bibitem{AnatoBrain}
De Brebisson, A. and Montana, G., 2015. Deep neural networks for anatomical brain segmentation. In Proceedings of the IEEE CVPR Workshops (pp. 20-28).
 
\bibitem{AlexNet}
Krizhevsky, A., et al., 2012. Imagenet classification with deep convolutional neural networks. In NIPS (pp. 1097-1105).

\bibitem{CAE}
Masci, J., et al., 2011. Stacked convolutional auto-encoders for hierarchical feature extraction. In ICANN (pp. 52-59). 

\bibitem{Unet}
Ronneberger, O., et al., 2015. U-net: Convolutional networks for biomedical image segmentation. In MICCAI (pp. 234-241).

\bibitem{IS}
Ny´ul LG, Udupa JK, Zhang X. New variants of a method of MRI scale standardization. IEEE TMI. 2000; 19(2):143–50.

\bibitem{M-net} Mehta, R., and Sivaswamy, J., 2017. M-net: A Convolutional Neural Network for deep brain structure segmentation. In Biomedical Imaging (ISBI 2017), 2017 IEEE 14th International Symposium on (pp. 437-440). IEEE. 

\end{thebibliography}
\end{document}